# A real-time battle situation intelligent awareness system based on Meta-learning & RNN

Yuchun Li, Zihan Lin, Xize Wang, Chunyang Liu, Liaoyuan Wu, Fang Zhang *
School of Chemical and Pharmaceutical Engineering, Changsha University of Science and Technology, Changsha 410114,China

**Abstract**: In modern warfare, real-time and accurate battle situation analysis is crucial for making strategic and tactical decisions. The proposed real-time battle situation intelligent awareness system (BSIAS) aims at meta-learning analysis and stepwise RNN (recurrent neural network) modeling, where the former carries out the basic processing and analysis of battlefield data, which includes multi-steps such as data cleansing, data fusion, data mining and continuously updates, and the latter optimizes the battlefield modeling by stepwise capturing the temporal dependencies of data set. BSIAS can predict the possible movement from any side of the fence and attack routes by taking a simulated battle as an example, which can be an intelligent support platform for commanders to make scientific decisions during wartime. This work delivers the potential application of integrated BSIAS in the field of battlefield command & analysis engineering.

## 1. Introduction

As an important part of modern war, battlefield situation intelligent perception technology plays a crucial role in enhancing operational effectiveness and decision-making quality. In the context of informationized war, the complexity and uncertainty of the battlefield environment are increasing, and the traditional battlefield situational awareness methods have been difficult to meet the needs of modern war (*1, 2*). Therefore, exploring new battlefield situational awareness technologies to improve the accuracy and real-time perception has become a hot spot in current research. Intelligent battlefield situational awareness technology mainly researches how to effectively acquire, process and parse battlefield information to provide a scientific basis for operational decision-making. With the rapid development of science and technology, the battlefield environment has become more and more complex, and the accuracy and real-time requirements for battlefield situational awareness have become higher and higher. Traditional battlefield situational awareness methods, such as manual analysis and simple model prediction, have been difficult to meet the needs of modern warfare (*3, 4*). These methods often suffer from low computational efficiency and poor prediction accuracy when dealing with large-scale, high-dimensional and dynamically changing battlefield data. Therefore, it is of great practical significance to study new

* Corresponding author, E-mail address: zhangf@csust.edu.cn (F. Zhang).

battlefield situational awareness techniques to improve the accuracy and real-time perception.

The research on battlefield situational awareness technology is relatively not very mature. The application of artificial intelligence techniques such as meta-learning and recurrent neural networks (RNN) in battlefield situational awareness has not been widely studied. Searching with meta-analysis and RNN model as keywords, there are 122 papers from 2014-01 to 2024-12, and the average annual number of papers in the literature is 14. The top 30 journals in terms of publications are shown on the left side of Fig. 1, where the journal with the most publications is arXiv (Cornell University) (12); Research Square (Research Square) is in the second place with 5 publications; Concurrency and Computation: Practice and Experience is in the third place with 5 publications; and Concurrency and Computation Practice and Experience is in third place with 3 publications. The annual publication volume is shown on the right side of Fig. 1, with 2023 reaching a peak of 41 annual publications and the fastest growth rate of 200% in 2017, suggesting that research in this field has been developing rapidly and is in a fast-rising phase (*5*).

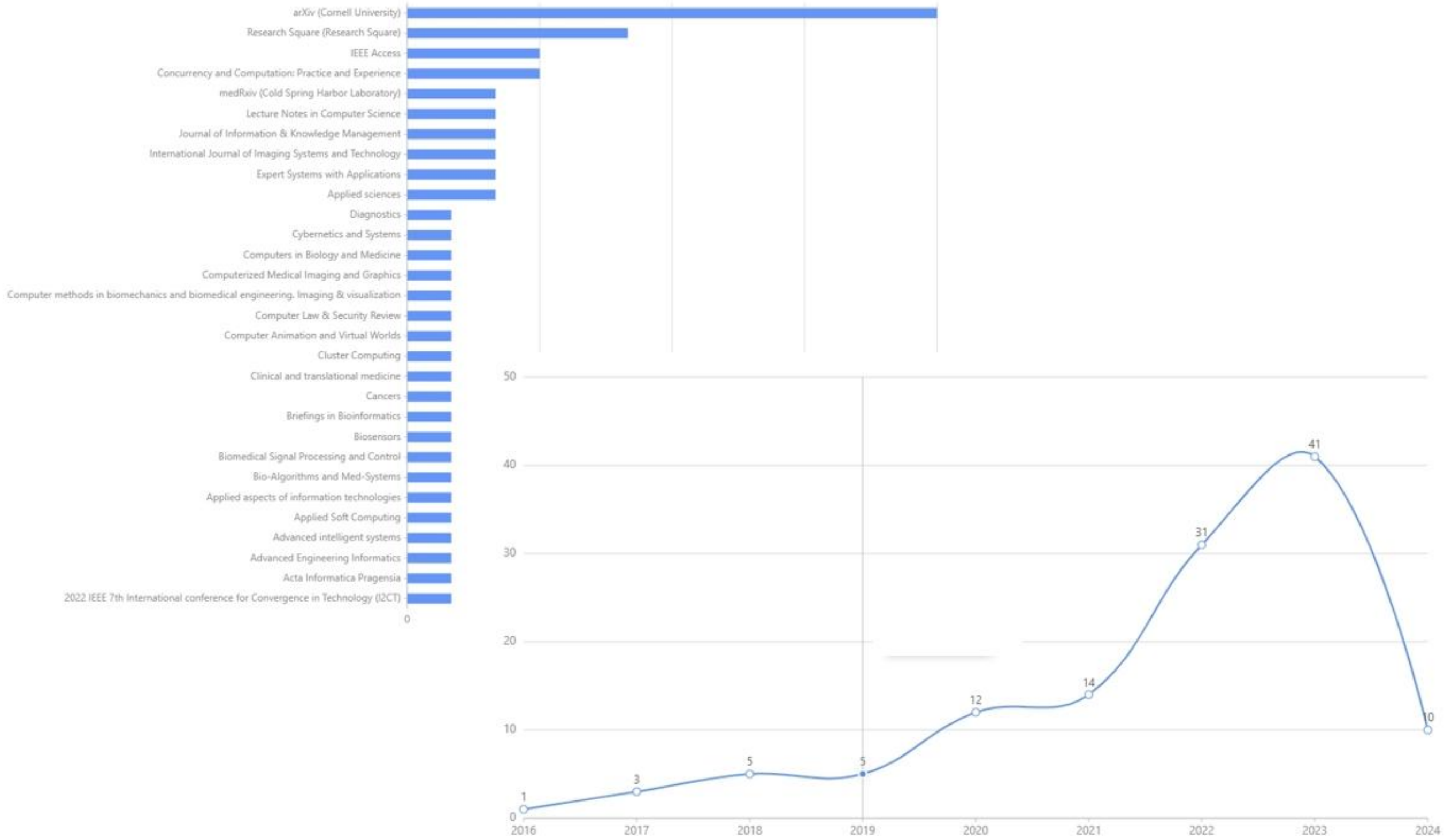

Fig. 1. Bibliographic analysis for publications and annual publication volume.

Meta-learning and RNN, as important techniques in the field of artificial intelligence, have shown great potential in battlefield situational awareness in recent years. Meta-learning is a method to improve learning efficiency by learning how to learn, which can migrate knowledge between different tasks, thus improving the generalization ability of battlefield situational awareness. On the other hand, RNN specialize in processing sequential data and can capture the time-dependency of battlefield data to improve the real-time nature of perception. Combining meta-learning and RNN can give full play to their respective advantages and improve the intelligence of battlefield situational awareness (*6, 7*). From 2014-01 to 2024-12, the top 20 national research institutions in the world in terms of publications in the field of meta-analysis and RNN model research are shown on the left side of Fig. 2, in which Duke University and Tsinghua University occupy the top two places in terms of publications, with 4 and 3 publications, respectively, and King Faisal University published 2 articles and is in the third place. The top 41 countries/regions are shown on the right side of Fig. 2. China (31 articles, 25.41%), India (30 articles, 24.59%), and United States of America (24 articles, 19.67%) are the second and the third countries/regions with the highest number of articles in this field (*5*).

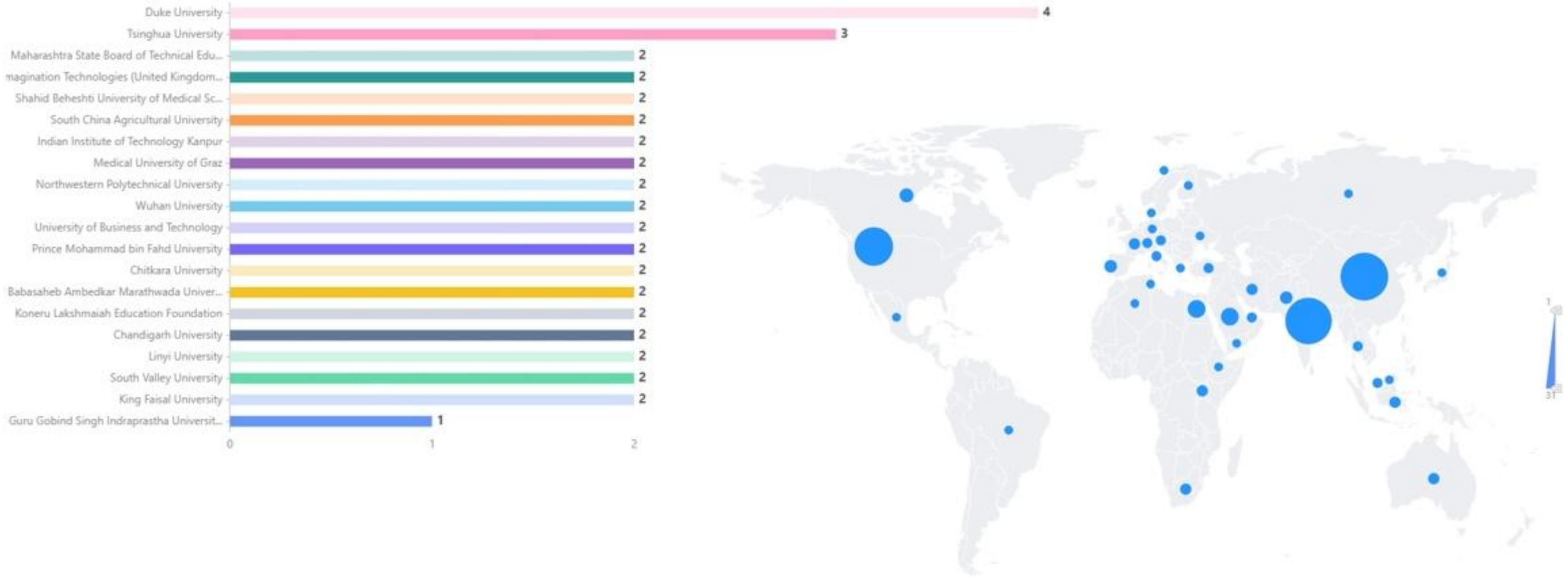

Fig. 2. Bibliographic analysis for publications and annual publication volume.

J. Dong reported unmanned aerial vehicles (UAVs) and unmanned combat vehicles to build a mobile ad hoc network that meets the conditions of the battlefield (*8*). G. R. Zibetti discussed IoT applications for decision-making process by delivering information collected directly from the battlefield to Command, Control, Communications, Computers, Intelligence, Surveillance and Reconnaissance (C4ISR) systems (*9, 10*). These systems can realize real-time collection, processing and transmission of battlefield information, providing more comprehensive, accurate and timely battlefield situational information for combat.

This study aims to combine the technical advantages of meta-learning and RNN to enhance the intelligent level of battlefield situational awareness with the support of R 4.4.2 software (*11*). This is not only of great significance for improving the accuracy and real-time performance of battlefield situational awareness, but also can provide technical reference and idea inspiration for other fields. Through BSIAS, we can better understand the application of meta-learning and RNN in battlefield situational awareness, and provide a useful reference for the future development of battlefield situational awareness technology. This work can provide a new methodology for the battle situation perception and command decision engineering.

## 2. Methodology of BSIAS

2.1. Presumption

Unlike data like the COVID-19 pandemic, which is continuously updated, information is very difficult to obtain during wartime, and even when data from a variety of sources is obtained, it is much less credible and reliable. However, even in the face of sparse or even NA source data, it is no longer too difficult to mine the true state of battlefield information under the current information technology with deep learning background. Before proposing the BSIAS, two assumptions need to be made at the outset, one is that although battlefield data is always non-exactly true for a variety of reasons, it always fluctuates around the true value rather than being completely out of line, and the other is that any change in the battlefield is based on a hyper multivariate, neural network type of structural framework as shown in Fig. 3.

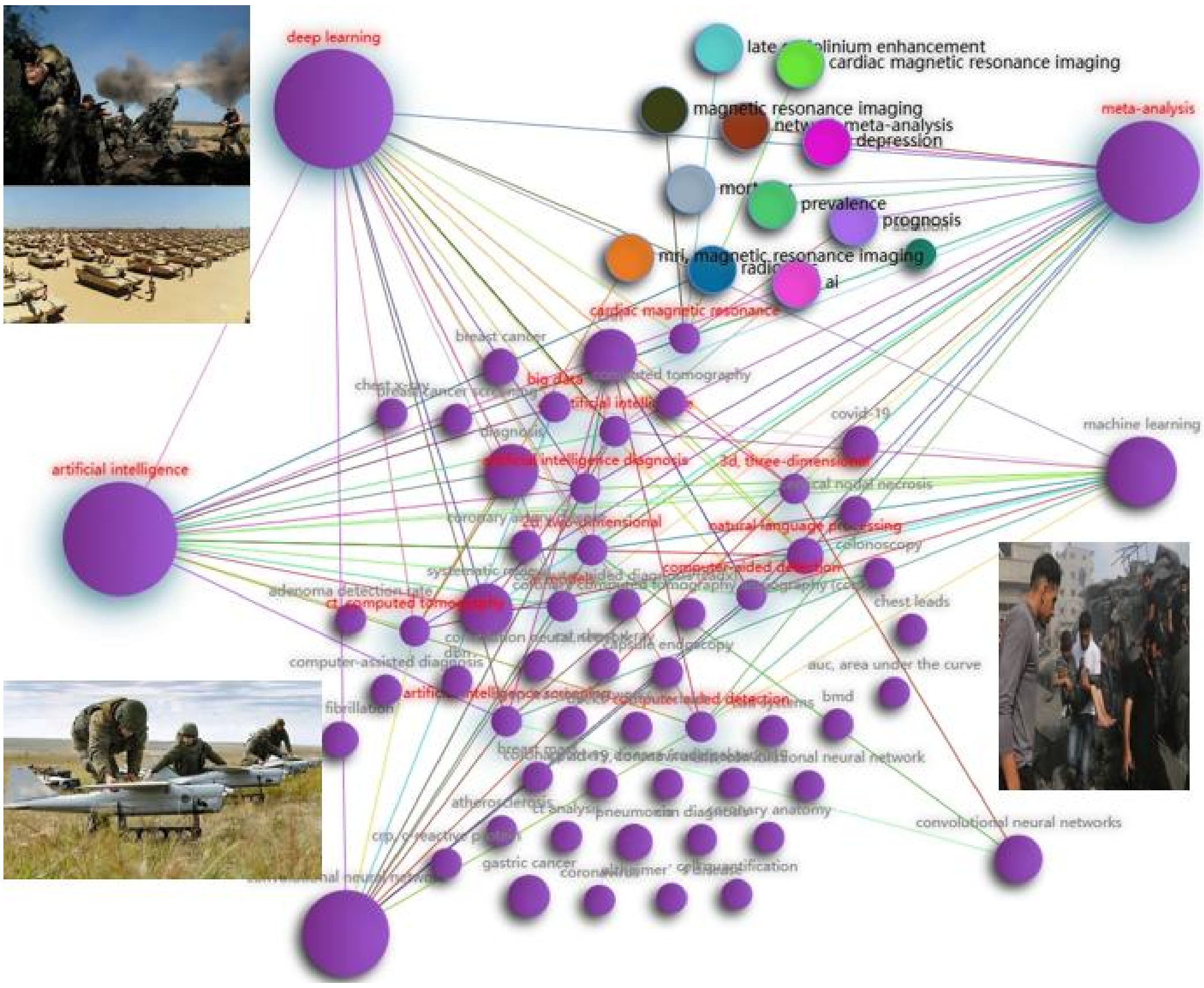

Fig. 3. Scheme of hyper multivariate, neural network type of structural framework.

2.2. Preliminary framework of BSIAS

BSIAS is constructed through statistical approaches and utilized for predicting whether the battle continued or not in the given input feature or data. Meta-learning analyses are mainly utilized in battle situation perception. The dataset is created with input data for predicting whether the battle is operating normally or not. The input characteristics of the battle contain metrics like finance, materials, population, territory and water resources. Fig. 4 displays main framework structure of battle situation, in which ”population” part is detailed discussed. According to the input characteristics, the battle deterioration or not is denoted in the binary target variable. Based on the input data metrics, the target classes are balanced for predicting the developments in warfare automatically. The deep learning (DL) techniques work effectively in the balanced datasets when compared to other unbalanced datasets. The predictions of the battle developments are complicated for determining the input data combination that leads to a deterioration result. A reliable and robust algorithm based on meta-learning and RNN is proposed for solving this type of scenario, as shown in Fig. 5.

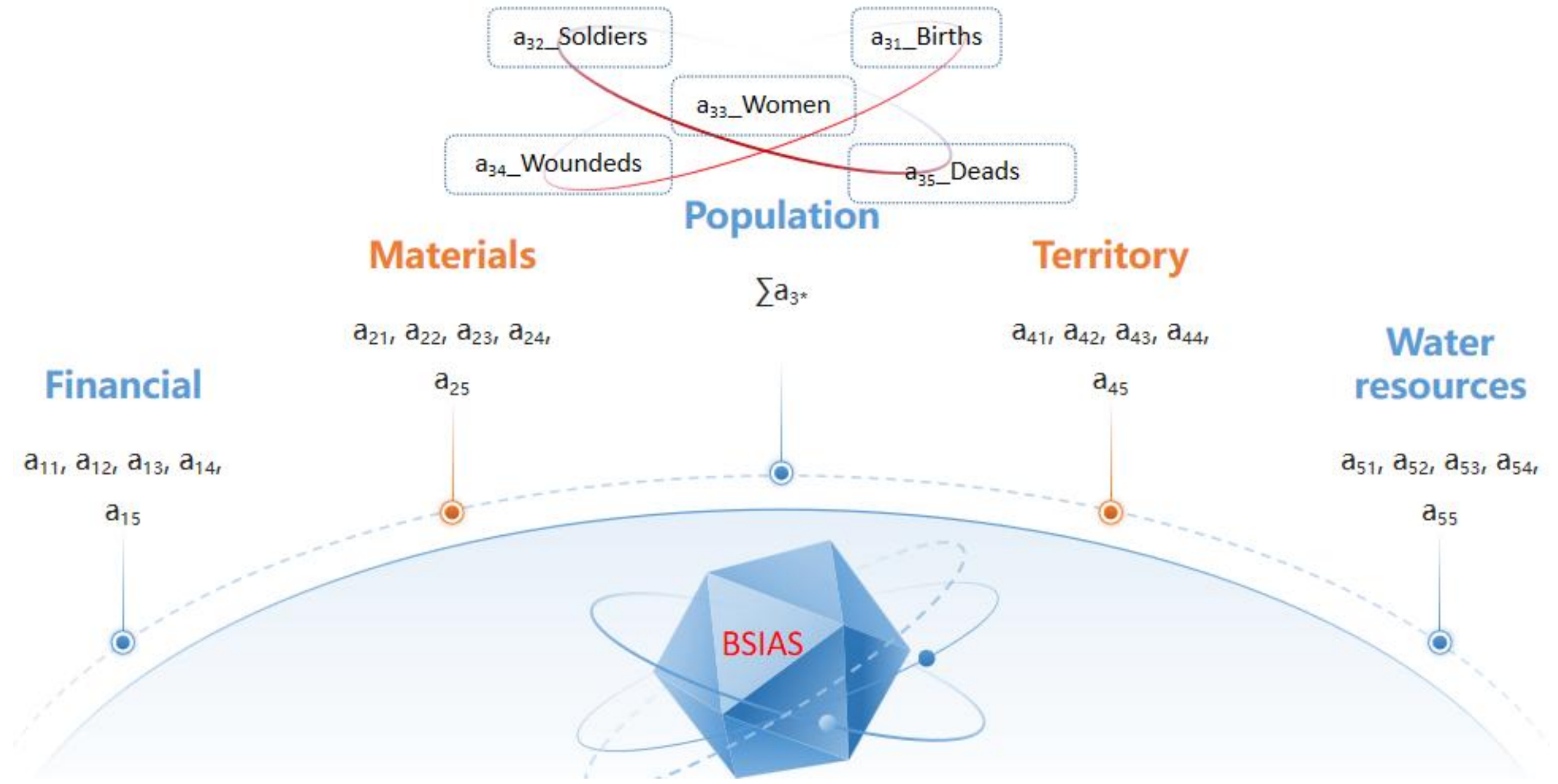

Fig. 4. Framework structure of BSIAS.

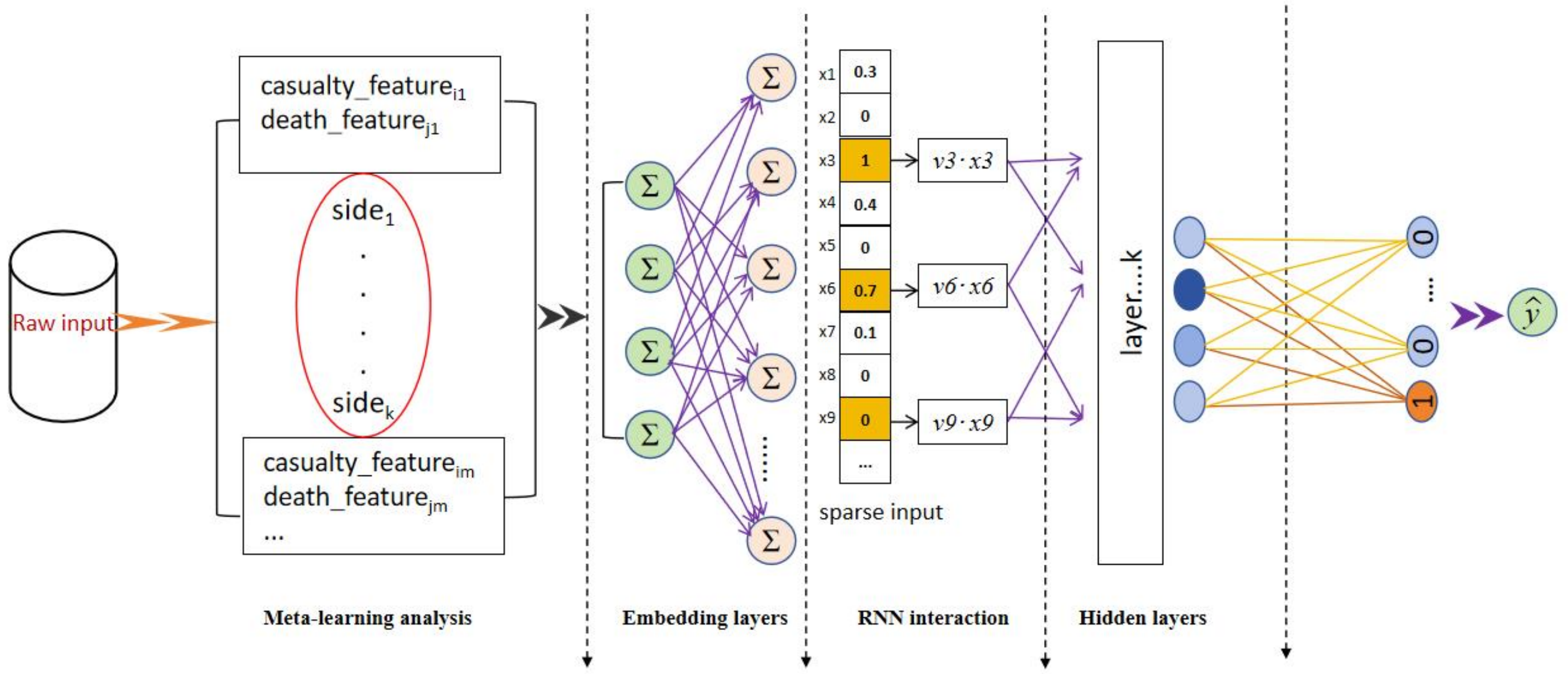

Fig. 5. Modeling scheme of BSIAS.

## 3. Experimental analysis

### 3.1. Meta-learning analysis of BSIAS

The parameter matrix consists of five first-level factors, namely financial, materials, population, territory and water resources, which can be referred to by the symbol $a_{1*}$, $a_{2*}$, $a_{3*}$, $a_{4*}$ and $a_{5*}$ correspondingly; each first-level factor consists of five second-level factors, which can be referred to by the symbol $a_{*1}$,...,$a_{*5}$. BSIAS can be described as $b_i + \sum \omega_j \bullet a_{mn}$, in which the relevant parameters are corrected of by the following-step feedback matrix. The unit process of meta-analysis is illustrated here using $a_{3*}$ as an example shown in Fig. 6, in which $a_{31}$, $a_{32}$, $a_{33}$, $a_{34}$ and $a_{35}$ refer to the number of births, the number of soldiers, the number of women, the number of wounded and the number of war dead in the population, respectively.

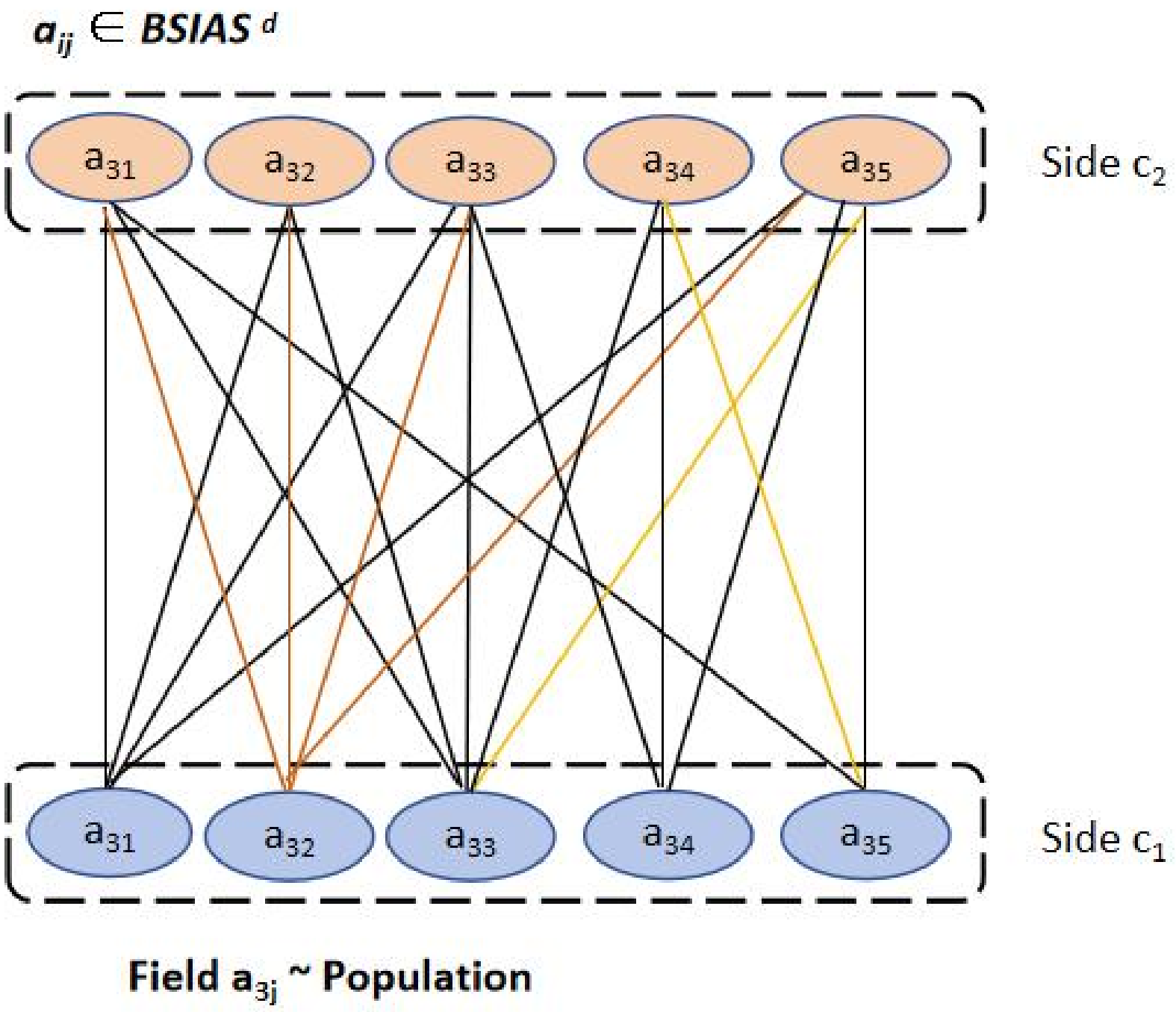

Fig. 6. Part parameter matrix constitution of BSIAS.

Assume a set of tuples $a_3(x_i, y)$, where $x_i \in a_{ij}$ is a feature vector and the corresponding target is $y$. For a34 and a35, Fig. 7 shows the structure of the two-arm meta-analysis, in which the hypothetical data source of casualties is located in the lower left corner of the diagram; The top half of the figure shows the forest plot for the two-arm analysis, in which weight vector of common and random are calculated and compared, and results show that heterogeneity of casualty data reported in the media from different sources was not significant.

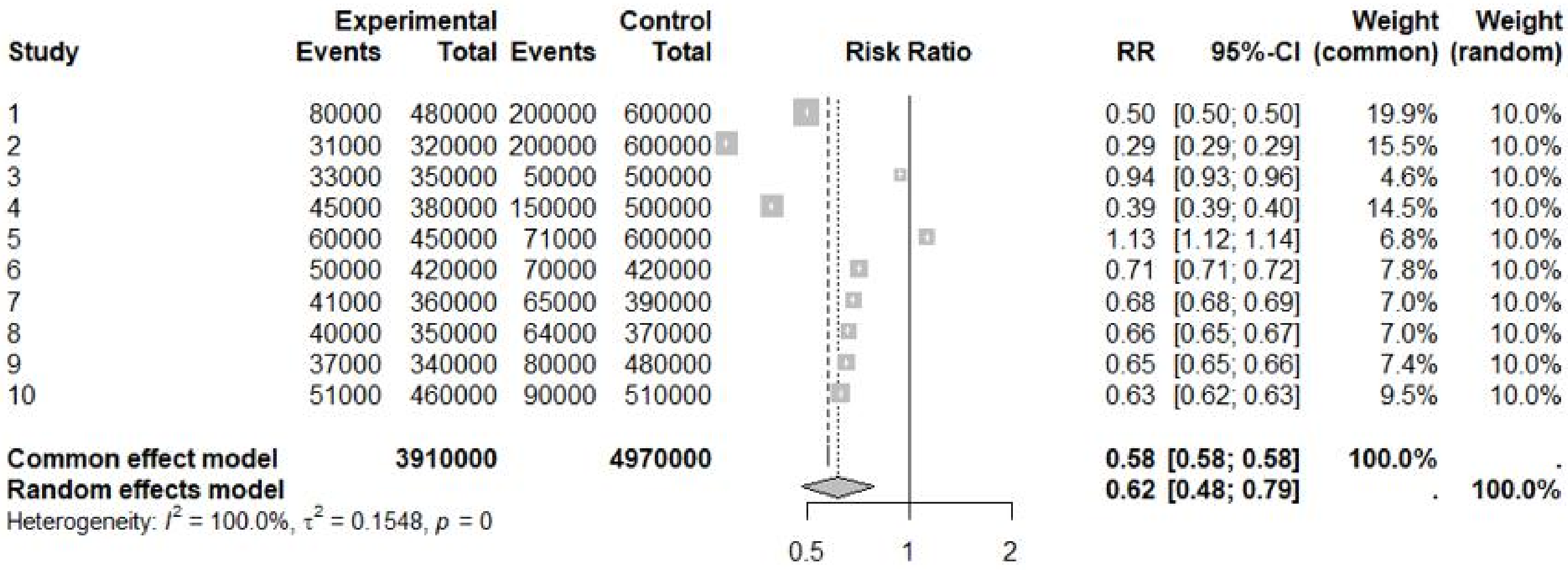

| Media_source | a34 | a35 | a34_2 | a35_2 |
|---|---|---|---|---|
| Wall Street Journal, 2024 | 480000 | 80000 | 600000 | 200000 |
| Ukrainian officials,2024 | 320000 | 31000 | 600000 | 200000 |
| UK Broadcasting, 2023 | 350000 | 33000 | 500000 | 50000 |
| France officals, 2024 | 380000 | 45000 | 500000 | 150000 |
| Mediazona, 2024 | 450000 | 60000 | 600000 | 71000 |
| Media_z1, 2023 | 420000 | 50000 | 420000 | 70000 |
| Media_z2, 2024 | 360000 | 41000 | 390000 | 65000 |
| Media_z3, 2023 | 350000 | 40000 | 370000 | 64000 |
| Media_z4, 2024 | 340000 | 37000 | 480000 | 80000 |
| Media_z5, 2024 | 460000 | 51000 | 510000 | 90000 |

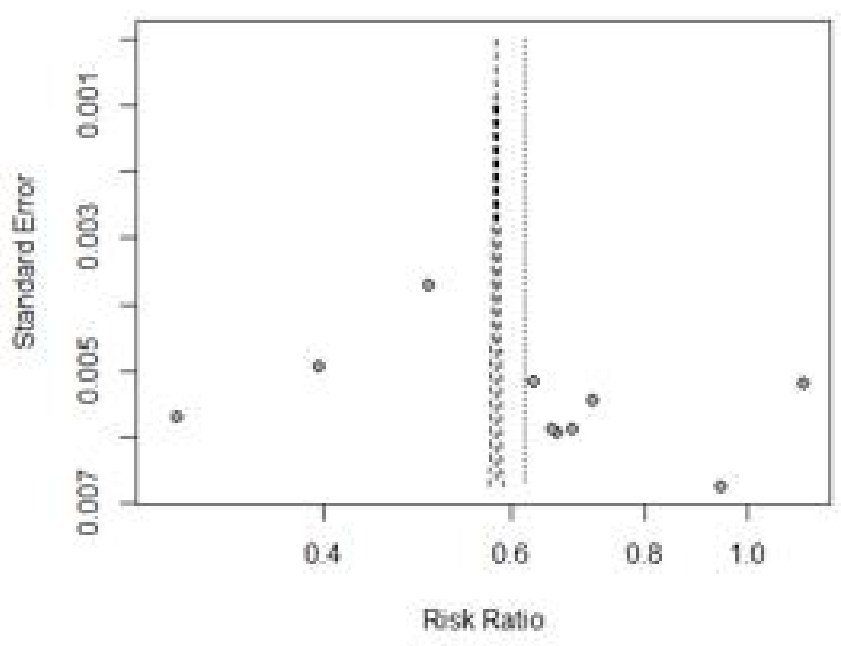

Fig. 7. Two-arm meta-analysis for a34, a35 ~ BSIAS.

Meta-learning is the strategy utilized to fuse the key characteristics with two or more base learners. The meta minimized the variance by predicting the errors because the framework is more robust than other designs, which can be used for assigning the pre-defined classifier weight for calculating. Based on the war casualty data from a single source, combined with the traditional rate of bias of that source and the number of sources it reports on (i.e., data such as learning rates are added), a comprehensive meta-analysis can be conducted to obtain the analytical framework shown in Fig.8. The forest plot, funnel plot and standard residual plot of the meta-analysis are given in the upper part of Fig. 8, and the statistical parameters characterizing the model are shown in the lower left part, containing the significance levels and probability values, with some of the input data (hypothetical data) shown in the lower right part of this figure.

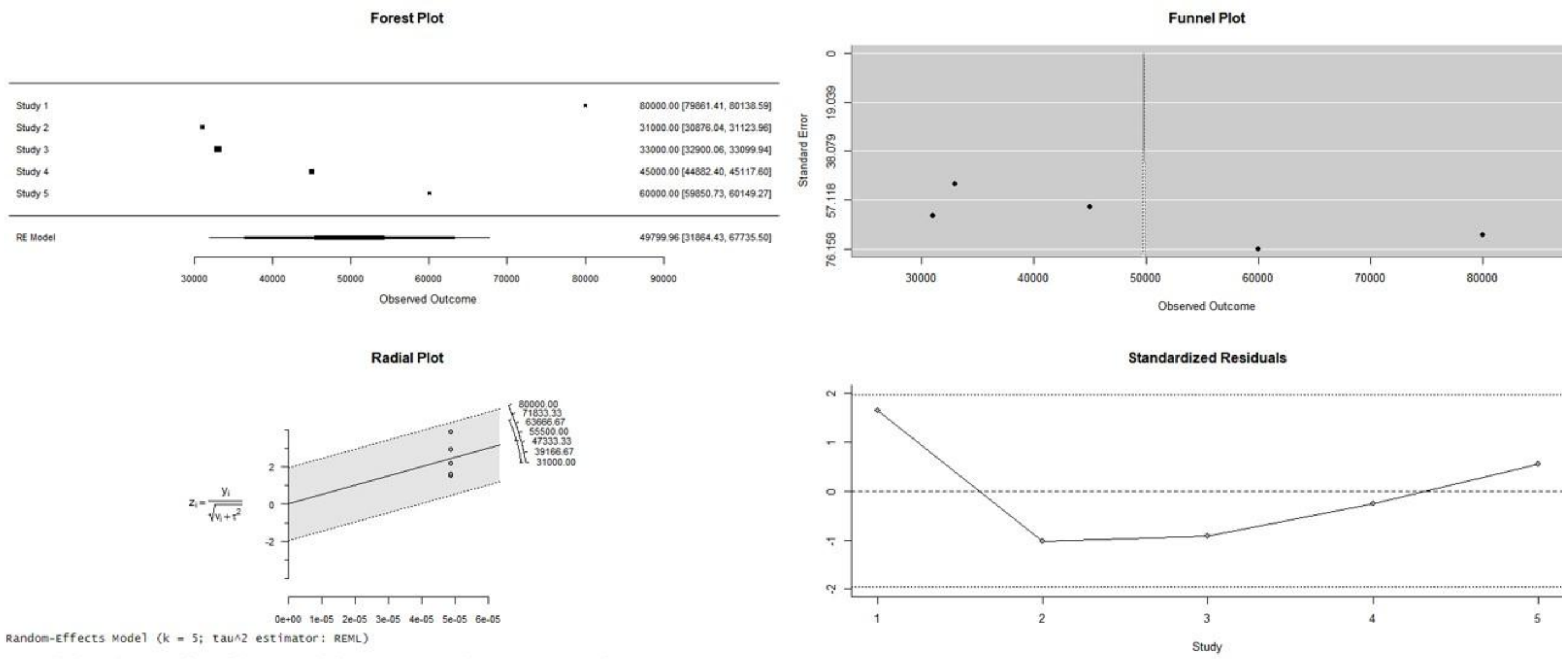

```
Random-Effects Model (k = 5; tau^2 estimator: REML)

tau^2 (estimated amount of total heterogeneity): 418695370.0974 (SE = 296065305.2934)
tau (square root of estimated tau^2 value):      20462.0471
I^2 (total heterogeneity / total variability):   100.00%
H^2 (total variability / sampling variability):  105375.97

Test for Heterogeneity:
Q(df = 4) = 386493.9498, p-val < .0001

Model Results:

  estimate        se     zval    pval       ci.lb       ci.ub
49799.9647  9150.9515  5.4421  <.0001  31864.4292  67735.5001  ***

---
Signif. codes:  0 '***' 0.001 '**' 0.01 '*' 0.05 '.' 0.1 ' ' 1
```

| Media_source | Media_id | a35 | a35_se | a35_n |
|---|---|---|---|---|
| Wall Street Journal, 2024 | 1 | 80000 | 5000 | 21 |
| Ukrainian officials,2024 | 2 | 31000 | 4000 | 19 |
| UK Broadcasting, 2023 | 3 | 33000 | 2600 | 24 |
| France officals, 2024 | 4 | 45000 | 3600 | 17 |
| Mediazona, 2024 | 5 | 60000 | 5800 | 9 |

Fig. 8. Comprehensive meta-analysis for a35 ~ BSIAS.

3.2. Stepwise RNN of BSIAS

Based on the hypothetical data set a34,a35 of Fig. 7, the virtual result column Y is added to obtain the original data set. In order to remove differences in numerical values and the influence of units among different variables, we need to apply a normalization function to scale each variable in the input data set, which maps the data uniformly onto the [0,1] interval for all variable. RNN modeling is performed for normalized data sets to obtain the model combination as shown in Fig. 9. The three network diagrams in Fig. 9 correspond to the RNN model with different hidden layers of 1, 2 and 3. Table 1 displays test errors and steps of each RNN model.

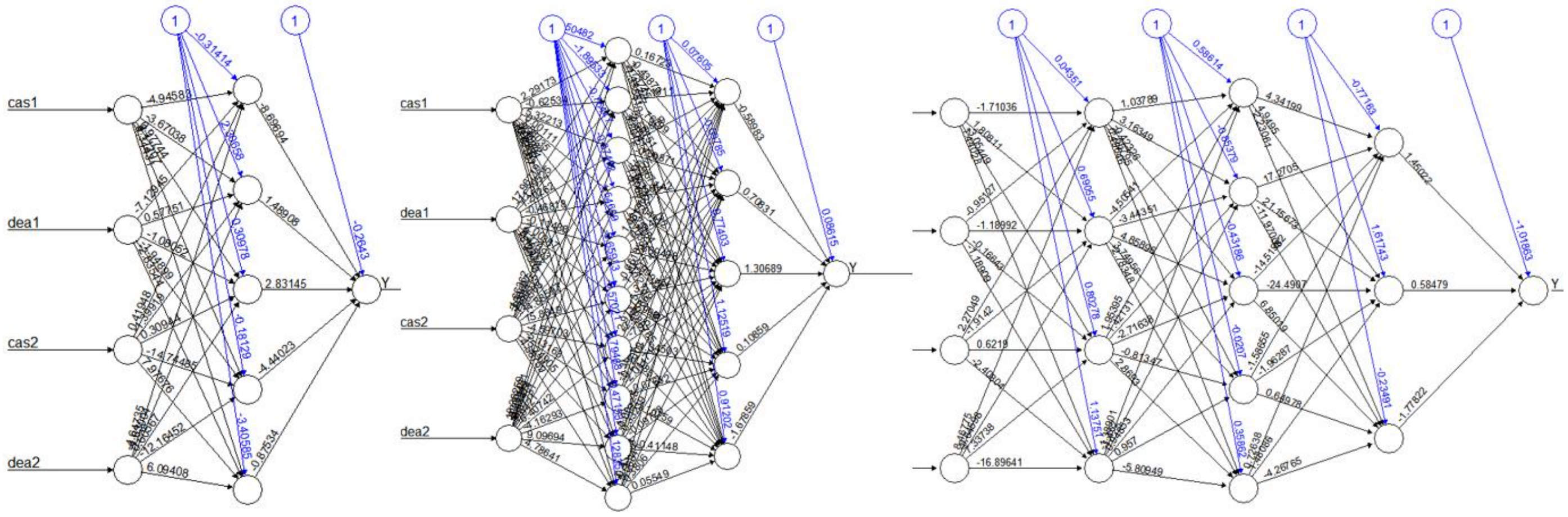

Fig. 9. Three RNN models with different hidden layers.

Table 1

Test errors and steps of RNN models through optimization.

| Model | Number of hidden layers | Number of neural cell for each layer | Errors | Steps |
|---|---|---|---|---|
| RNN_1 | 1 | 5 | 0.0043 | 512 |
| RNN_2 | 2 | 10, 5 | 0.0011 | 225 |
| RNN_3 | 3 | 4, 5, 3 | 0.0087 | 864 |

As can be seen from Table 1, the model with 2 hidden layers not only has the relatively smallest deviation value, but also requires only 225 steps for modelling, which is also the lowest of the three models. So two hidden layers are chosen as the main parameters for RNN modelling and the best number of neurons can be obtained by optimizing layer by layer. Accurate and reliable analytical models can be obtained through a stepwise RNN modelling approach, as shown in Fig. 10. In this figure, the stepwise RNN model provides the slope, intercept parameters of each neuron of the two hidden layers and also gives the weight list of the most dominant weight list i.e. the generalized weight matrix.

```
                    Length Class      Mode
call                 5     -none-     call
response            10     -none-     numeric
covariate           40     -none-     numeric
model.list           2     -none-     list
err.fct              1     -none-     function
act.fct              1     -none-     function
linear.output        1     -none-     logical
data                 5     data.frame list
exclude              0     -none-     NULL
net.result           1     -none-     list
weights              1     -none-     list
generalized.weights  1     -none-     list
startweights         1     -none-     list
result.matrix      114     -none-     numeric
```

| $result.matrix | |
|---|---|
| error | 0.001108828 |
| reached.threshold | 0.007565109 |
| steps | 225 |
| Intercept.to.1layhid1 | 1.284519361 |
| cas1.to.1layhid1 | -5.774945637 |
| dea1.to.1layhid1 | -3.238410577 |
| cas2.to.1layhid1 | 16.09171583 |
| dea2.to.1layhid1 | -0.68750789 |
| Intercept.to.1layhid2 | 1.022877981 |
| cas1.to.1layhid2 | -2.205639876 |
| dea1.to.1layhid2 | 0.459555627 |
| ... | ... |

```
$generalized.weights
$generalized.weights[[1]]
            [,1]         [,2]        [,3]       [,4]
 [1,]    1.911045    4.7950330   18.221344  -29.53282
 [2,]   15.894648   29.6627056  -24.542049   27.05209
 [3,] -120.277735  164.1584806 -123.233296  121.48759
 [4,]    6.265393   -0.6616818   -9.247488   15.89105
 [5,]  -98.294899  195.0596758  -66.677910  515.04840
 [6,] -752.173623 -104.7000986 1431.611317  348.40173
 [7,] -113.683884  -10.2742018  207.693408   48.34901
 [8,] -542.843496 -111.7472507 1247.503667  275.50731
 [9,] -274.585347 -963.9406510 1005.090564 -736.70162
[10,]   43.821162  -21.3406118  -32.169789  -82.50436
```

Fig. 10. Results and weights matrix of model with two hidden layers.

## 4. Conclusions

A real-time battle situation intelligent awareness system was proposed through meta-learning analysis and stepwise RNN (recurrent neural network) modeling. Meta-learning analysis carries out the basic processing and analysis of battlefield data, which includes multi-steps such as data cleansing, data fusion, data mining and continuously updates, and the stepwise RNN optimizes the battlefield modeling by stepwise capturing the temporal dependencies of data set. BSIAS can better understand the application of meta-learning and stepwise RNN in battlefield situational awareness, and provide a useful reference for the future development of battlefield situational awareness technology. This work delivers the potential application of integrated BSIAS in battlefield command & analysis engineering.

## Acknowledgments

The authors thank the support from Hunan Provincial Key Laboratory of Materials Protection for Electric Power and Transportation (Changsha University of Science & Technology). We also appreciate the strong support of Zixin Li from Changsha Power Supply Branch Company, State Grid Hunan Electric Power Co..

## CRediT authorship contribution statement

Yuchun Li: Writing – original draft. Zihan Lin: Methodology. Xize Wang: Investigation. Chunyang Liu: Formal analysis. Liaoyuan Wu: Writing – review & editing. Fang Zhang: Supervision.

## Declaration of Competing Interest

The authors declare that they have no known competing financial interests or personal relationships that could have appeared to influence the work reported in this paper.

## Data availability

Data will be made available on request.